\pdfoutput=1

\documentclass[12pt,a4paper]{article}
\usepackage[top=1in, bottom=1in, left=1in, right=1in]{geometry}

\usepackage{lastpage,fancyhdr,graphicx}
\usepackage{algorithm}
\usepackage{algpseudocode}

\usepackage{amsmath,amssymb}

\usepackage{changepage}

\usepackage[utf8x]{inputenc}

\usepackage{textcomp,marvosym}

\usepackage{cite}

\usepackage{nameref,hyperref}

\usepackage[right]{lineno}

\usepackage{microtype}
\DisableLigatures[f]{encoding = *, family = * }

\usepackage[table]{xcolor}

\usepackage{array}

\newcolumntype{+}{!{\vrule width 2pt}}

\newlength\savedwidth

\newcommand*{\bigcdot}{\raisebox{-0.25ex}{\scalebox{1.2}{$\cdot$}}}

\usepackage[aboveskip=1pt,labelfont=bf,labelsep=period,justification=raggedright,singlelinecheck=off]{caption}

\makeatletter
\renewcommand{\@biblabel}[1]{\quad#1.}
\makeatother

\date{}

\begin{document}

\title{Neuron's Eye View: Inferring Features of Complex Stimuli from Neural Responses}
\author{Xin (Cindy) Chen\textsuperscript{1,3}, Jeffrey M. Beck\textsuperscript{2,3}, John M. Pearson\textsuperscript{1,3*}}
\maketitle

\noindent
\textbf{1} Duke Institute for Brain Sciences, Duke University, Durham, North Carolina, USA
\\
\textbf{2} Department of Neurobiology, Duke University Medical Center, Durham, North Carolina, USA
\\
\textbf{3} Center for Cognitive Neuroscience, Duke University, Durham, North Carolina, USA

%
%


\noindent
* john.pearson@duke.edu

\section*{Abstract}
Experiments that study neural encoding of stimuli at the level of individual neurons typically choose a small set of features present in the world --- contrast and luminance for vision, pitch and intensity for sound --- and assemble a stimulus set that systematically varies along these dimensions. Subsequent analysis of neural responses to these stimuli typically focuses on regression models, with experimenter-controlled features as predictors and spike counts or firing rates as responses. Unfortunately, this approach requires knowledge in advance about the relevant features coded by a given population of neurons. For domains as complex as social interaction or natural movement, however, the relevant feature space is poorly understood, and an arbitrary \emph{a priori} choice of features may give rise to confirmation bias. Here, we present a Bayesian model for exploratory data analysis that is capable of automatically identifying the features present in unstructured stimuli based solely on neuronal responses. Our approach is unique within the class of latent state space models of neural activity in that it assumes that firing rates of neurons are sensitive to multiple discrete time-varying features tied to the \emph{stimulus}, each of which has Markov (or semi-Markov) dynamics. That is, we are modeling neural activity as driven by multiple simultaneous stimulus features rather than intrinsic neural dynamics.  We derive a fast variational Bayesian inference algorithm and show that it correctly recovers hidden features in synthetic data, as well as ground-truth stimulus features in a prototypical neural dataset. To demonstrate the utility of the algorithm, we also apply it to cluster neural responses and demonstrate successful recovery of features corresponding to monkeys and faces in the image set.

\section*{Introduction}
The question of how the brain encodes information from the natural world forms one of the primary areas of study within neuroscience. For many sensory systems, particularly vision and audition, the discovery that single neurons modulate their firing of action potentials in response to particular stimulus features has proven foundational for theories of sensory function. Indeed, neuronal responses to contrast, edges, and motion direction appear to form fundamental primitives on which higher-level visual abstractions are built. Nevertheless, many of these higher-level abstractions do not exist in a stimulus space with obvious axes. As a result, experimenters must choose \emph{a priori} features of interest in constructing their stimulus sets, with the result that cells may appear weakly tuned due to misalignment of stimulus and neural axes.

For example, in vision, methods like reverse correlation have proven successful in elucidating response properties of some cell types, but such techniques rely on a well-behaved stimulus space and a highly constrained encoding model in order to achieve sufficient statistical power to perform inference \cite{steveninck1988realtime,ringach2004reverse,ringach2002receptive}. However, natural stimuli are known to violate both criteria, generating patterns of neural activity that differ markedly from those observed in controlled experiments with limited stimulus complexity \cite{ringach2002receptive,sharpee2004analyzing,Vinje2000-dx}. Information-based approaches have gone some way in addressing this challenge \cite{sharpee2004analyzing}, but this approach assumes a metric structure on stimuli in order to perform optimization, and was recently shown to be strongly related to standard Poisson regression models\cite{Williamson2013-rg}.

More recently, Gallant and collaborators have tackled this problem in the context of fMRI, demonstrating that information present in the blood oxygen level-dependent (BOLD) signal is sufficient to classify and map the representation of natural movie stimuli across the brain \cite{Vu2011-da,Huth2012-cj,Stansbury2013-nm}. These studies have used a number of modeling frameworks, from Latent Dirichlet Allocation for categorizing scene contents \cite{Stansbury2013-nm} to regularized linear regression \cite{Huth2012-cj} to sparse nonparametric models \cite{Vu2011-da} in characterizing brain encoding of stimuli, but in each case, models were built on pre-labeled training data. Clearly, a method that could infer stimulus structure directly from neural data themselves could extend such work to less easily characterized stimulus sets like those depicting social interactions.

Another recent line of work, this one focused on latent Poisson processes, has addressed the task of modeling the low dimensional dynamics of neural populations\cite{Pillow2008-em,Vogelstein2009-ax,Park2014-el,Buesing2014-ta,Archer2015-ec, Zhao2016-bw,Gao2016-ck}. Using generalized linear models and latent linear dynamical systems as building blocks, these models have proven able to infer (functional) connectivity \cite{Pillow2008-em}, estimate spike times from a calcium images\cite{Vogelstein2009-ax}, and identify subgroups of neurons that share response dynamics\cite{Buesing2014-ta,Zhao2016-bw,Gao2016-ck}. Inference in these models is generally performed via expectation maximization, though \cite{Ulrich2014-zc,Putzky2014-up,Archer2015-ec, Zhao2016-bw,Gao2016-ck} also used a variational Bayesian approach. Our work is distinct from those models, however, in that those were concerned with modeling and discriminating \emph{internal} states based on neural responses, while this work focuses on detecting features in \emph{external} stimuli. Moreover, in contrast to \cite{Buesing2014-ta,Archer2015-ec,Zhao2016-bw,Gao2016-ck}, we focus on multiple binary latent states as a means of ``tagging" a finite number of overlapping stimulus features.

Our model sits at the intersection of these regression and latent variable approaches. We utilize a Poisson observation model that shares many of the same features as the commonly used generalized linear models for Poisson regression. We also assume that the latent features modulating neural activity are time-varying and Markov. However, we make 3 additional unique assumptions: First, we assume that the activity of each neuron is modulated by a combination of multiple independent latent features governed by (semi-)Markov dynamics. This allows for latents to evolve over multiple timescales with non-trivial duration distributions, much like the hand-labeled features in social interaction data sets. Second, we assume that these latents are tied to stimulus presentation. That is, when identical stimuli are presented, the same latents are also present. This allows us to model the dynamics of latent features of the \emph{stimulus} that drive neural activity, rather than intrinsic neural dynamics. Finally, we enforce a sparse hierarchical prior on modulation strength that effectively limits the number of latent features to which the population of neurons is selective. This allows for a parsimonious explanation of the firing rates of single units in terms of a small set of stimulus features. Finally, we perform full variational Bayesian inference on all model parameters and take advantage of conditional conjugacy to generate coordinate ascent update rules, nearly all of which are explicit. Combined with forward-backward inference for latent states, our algorithm is exceptionally fast, automatically implements Occam's razor, and facilitates proper model comparisons using the variational lower bound.

\section*{Model}
\label{model_sec}

\subsection*{Observation model}
\begin{figure}[!ht]
    \includegraphics[width=\linewidth]{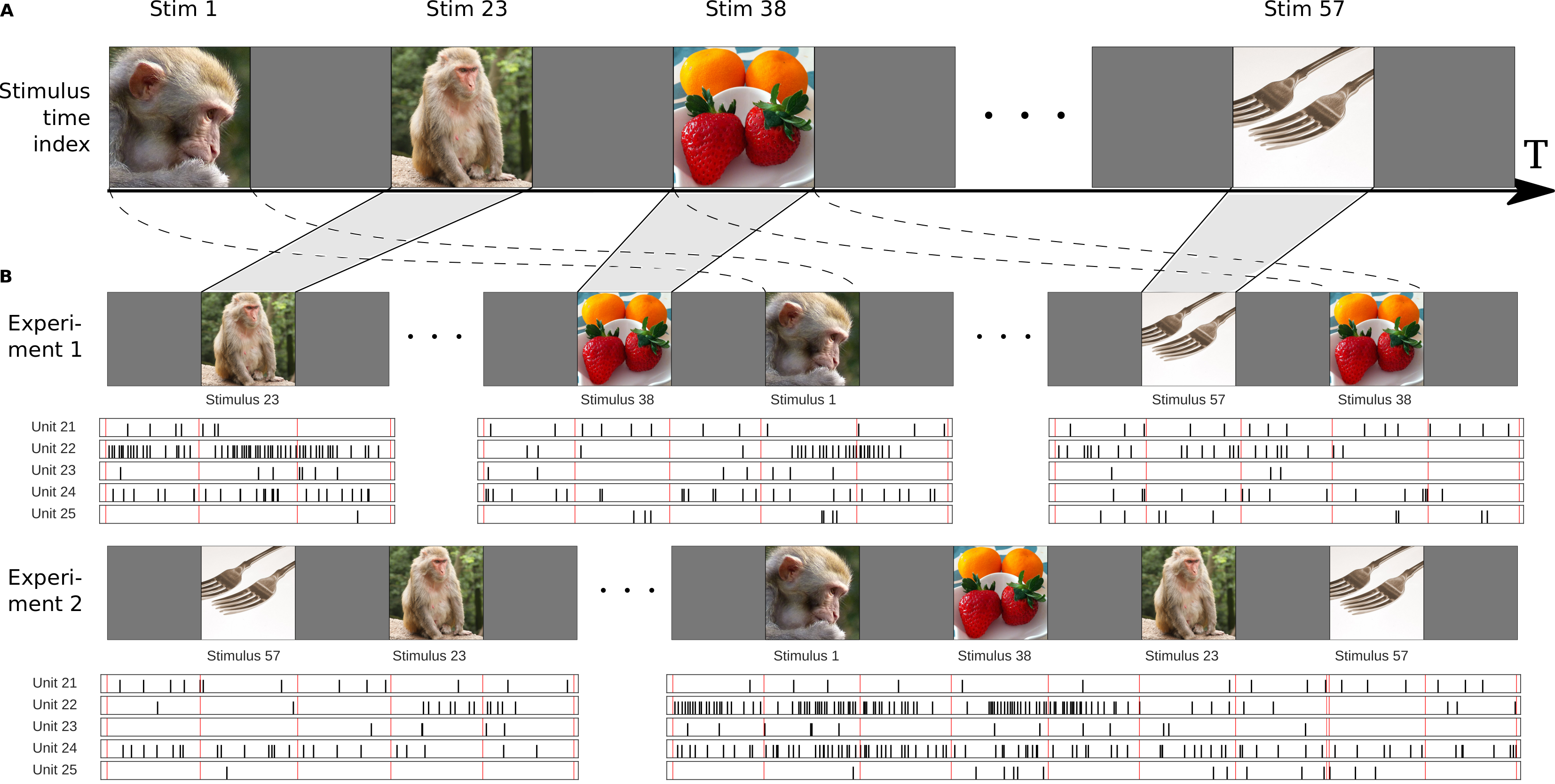}
	\caption{\bf Observational model.}
    A: Stimuli are concatenated to form a single time series indexed by $t$. B: Individual experimental sessions draw from the available set of stimuli, with index $m$ representing unique $(time, unit)$ presentations. Example stimulus sequences for two experiments are shown, with corresponding neuronal spike data. Note that the number of exposure times $M_{tu}$ for each stimulus and unit can be different.
	\label{fig:movie}
\end{figure}

Consider a population of $U$ spiking neurons or units exposed to a series of stimuli indexed by a time index $t\in \lbrace 1\ldots T\rbrace$. We assume that this time index is unique across all stimuli, such that a particular $t$ represents a unique moment in a particular stimulus. In order to model repeated presentations of the same stimulus to the same neuron, we further assume that each neuron is exposed to a stimulus $M_{tu}$ times, though we do not assume any relationship among $M_{tu}$. That is, we need not assume either that all neurons see each stimulus the same number of times, nor that each stimulus is seen by all neurons. It is thus typical, but not required, that $M_{tu}$ be sparse, containing many 0s, as shown in Figure \ref{fig:movie}.

For each observation $m$ in $M_{tu}$, we then observe a spike count, $N_m$. Note that $m$ is a unique $(time, unit)$ pair that can be denoted by $(t(m), u(m))$. We model these spike counts as arising from a Poisson process with time-dependent rate $\Lambda_{tu}$ and observation-specific multiplicative overdispersion $\theta_m$:
\begin{align}
    \label{obs_model}
    N_{m} &\sim \text{Pois}(\Lambda_{t(m), u(m)} \theta_m) &
    \text{where} ~~ \theta_m &\sim \text{Gamma}(s_{u(m)}, s_{u(m)})
\end{align}
That is, for a given stimulus presentation, the spiking response is governed by the firing rate $\Lambda$, specific to the stimulus and unit, along with a moment-by-moment noise in the unit's gain, $\theta_m$. We restrict these $\theta_m$ to follow a Gamma distribution with the same shape and rate parameters, since this results in an expected noise gain of 1. In practice, we model this noise as independent across observations, though it is possible to weaken this assumption, allowing for $\theta_m$ to be autocorrelated in time (see Supplement). Note that both the unit and time are functions of the observation index $m$, and that the distribution of the overdispersion for each observation may be specific to the unit observed.

\subsection*{Firing rate model}
At each stimulus time $t$, we assume the existence of $K$ binary latent states $z_{tk}$ and $R$ observed covariates $x_{tr}$. The binary latent states can be thought of as time-varying ``tags'' of each stimulus --- for example, content labels for movie frames --- and are modeled as Markov chains with initial state probabilities $\pi_k$ and transition matrices $A_k$. The observed covariates, by contrast, are known to the experimenter and may include contrast, motion energy, or any other \emph{a priori} variable of interest.

We further assume that each unit's firing rate at a particular point in time can be modeled as arising from the product of three effects: (1) a baseline firing rate specific to each unit ($\lambda_0$), (2) a product of responses to each latent state ($\lambda_z$), and (3) a product of responses to each observed covariate ($\lambda_x$):
\begin{equation}
    \label{fr_model}
    \Lambda_{tu} = \lambda_{0u} \prod_{k = 1}^K (\lambda_{zuk})^{z_{tk}}
    \prod_{r = 1}^R (\lambda_{xur})^{x_{tr}}
\end{equation}
Note that this is conceptually similar to the generalized linear model for firing rates (in which we model $\log \Lambda$) with the identification $\beta = \log \lambda$. However, by modeling the firing rate as a product and placing Gamma priors on the individual effects, we will be able to take advantage of closed-form variational updates resulting from conjugacy that avoid explicit optimization (see below). Note also, that because we assume the $z_{tk}$ are binary, the second term in the product above simply represents the cumulative product of the gain effects for those features present in the stimulus at a given moment in time.

In addition, to enforce parsimony in our feature inference, we place sparse hierarchical priors with hyperparameters $\gamma = (c, d)$ on the $\lambda_z$ terms:
\begin{align}
    \label{hierarchy}
    \lambda_{zuk} &\sim \text{Gamma}(c_{zk}, c_{zk} d_{zk}) & c_{zk} &\sim \text{Gamma}(a_{ck}, b_{ck})
    & d_{zk} &\sim \text{Gamma}(a_{dk}, b_{dk})
\end{align}
That is, the population distribution for the responses to latent features is a gamma distribution, with parameters that are themselves gamma-distributed random variables. As a result, $\mathbb{E}[\lambda_u] = d^{-1}$ and $\text{var}[\lambda_u] = (cd^2)^{-1}$, so in the special case of $c$ large and $d\sim \mathcal{O}(1)$, the prior for firing rate response to each latent feature will be strongly concentrated around gain 1 (no effect). As we show below, this particular choice results in a model that only infers features for which the data present strong evidence, controlling for spurious feature detection. In addition, this particular choice of priors leads to closed-form updates in our variational approximation. For the baseline terms, $\lambda_{0u}$, we use a non-sparse version of the same model; for the covariate responses, $\lambda_{xu}$, we model the unit effects non-hierarchically, using independent Gamma priors for each unit.

\begin{figure}[!ht]
    \includegraphics[width=\linewidth]{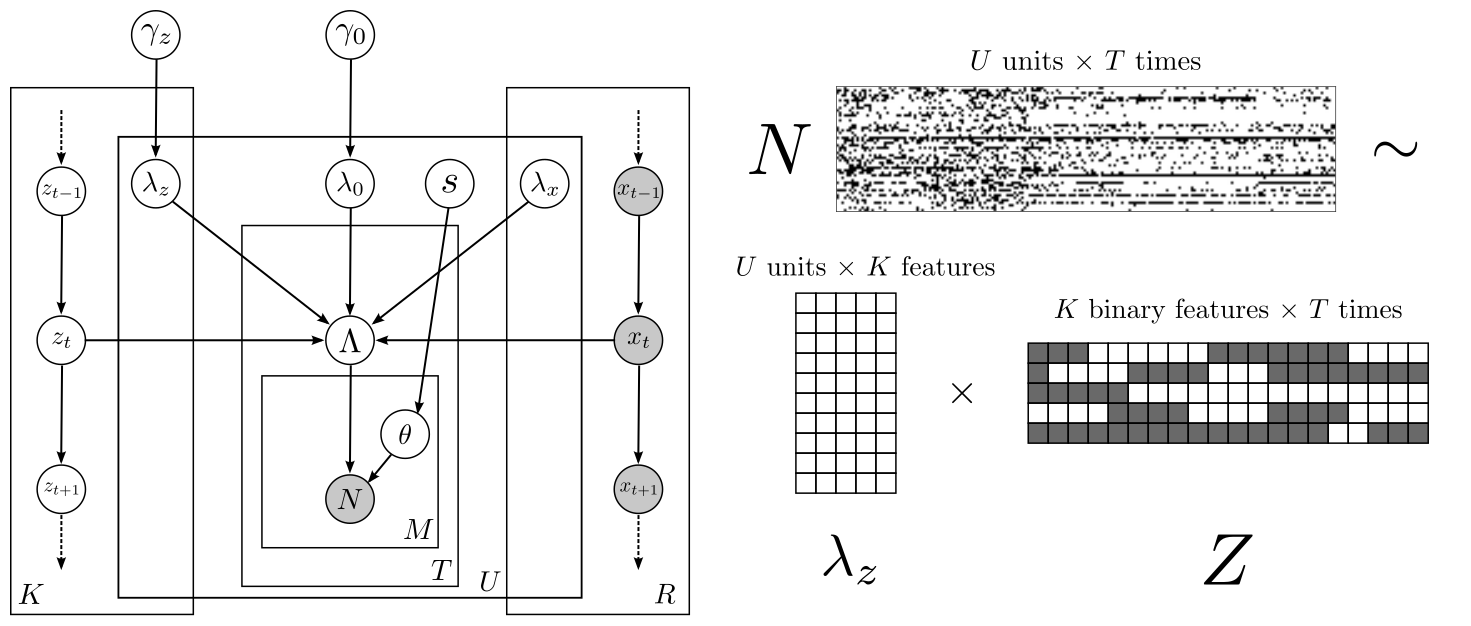}
	\caption{\bf Generative model for spike counts.}
	A: Counts are assumed Poisson-distributed, with firing rates $\Lambda$ that depend on each unit's responses ($\lambda$) to both latent discrete states $z_t$ and observed covariates $x_t$ that change in time, as well as a baseline firing rate $\lambda_0$. $\gamma$ nodes represent hyperparameters for the firing rate effects. $\theta$ is a multiplicative overdispersion term specific to each observation, distributed according to hyperparameters $s$. B: Spike counts $N$ are observed for each of $U$ units over stimulus time $T$ for multiple presentations $M_{tu}$.
\label{fig1}
\end{figure}

Putting all this together, we then arrive at the full generative model:
\begin{align}
    p(N, \Lambda, \theta) &= p(N| \Lambda, \theta)p(\Lambda|\lambda, z)
    p(\lambda|\gamma) p(\gamma)
    p(z|A, \pi)
    p(A)p(\pi)p(\theta|s)p(s) \\
    \text{where} ~~ p(\lambda|\gamma) &= \prod_u p(\lambda_{0u}|c_0, d_0)\prod_{kr} p(\lambda_{zuk}|c_{zk}, d_{zk}) p(\lambda_{xur}) \\
    \text{~~ and} ~~ p(\gamma) &= p(c_0)p(d_0)\prod_k p(c_{zk}) p(d_{zk})
\end{align}
in conjunction with the definitions of $p(N|\Lambda, \theta)$ and $\Lambda(\lambda, z, x)$ in Eq (\ref{obs_model}) and (\ref{fr_model}). The generative model for spike counts is illustrated in Figure \ref{fig1}.

\section*{Inference}
Given a sequence of stimulus presentations $(t(m),u(m))$ and observed spike counts $N_m$, we want to infer both the model parameters $\Theta = (\lambda_0, \lambda_z, \lambda_x, A, \pi, c_0, d_0, c_z, d_z, s)$ and latent variables  $Z=(z_{kt},\theta_m)$. That is, we wish to calculate the joint posterior density:
\begin{equation}
    p(\Theta,Z|N) \propto p(N | Z, \Theta) p(Z) p(\Theta)
\end{equation}
In general, calculating the normalization constant for this posterior is computationally intractable. Instead, we will use a variational approach, approximating $p(\Theta, Z|N)$ by a variational posterior $q(Z, \Theta) = q_Z(Z) q_\Theta(\Theta)$ that factorizes over parameters and latents but is nonetheless close to $p$ as measured by the Kullback-Leibler divergence \cite{Wainwright2008-ii}. Equivalently, we wish to maximize the variational objective
\begin{equation}
    \label{elbo}
    \mathcal{L} \equiv \mathbb{E}_q \left[\log \frac{p(\Theta,Z|N)}{q(\Theta, Z)} \right] = \mathbb{E}_q \left[\log p(\Theta,Z|N) \right] + \mathcal{H}[q_\Theta(\Theta)] + \mathcal{H}[q_Z(Z)]
\end{equation}
with $\mathcal{H}$ the entropy. We adopt the factorial HMM trick of \cite{ghahramani1997factorial}, making the reasonable assumption that the posterior factorizes over each latent time series $z_{\bigcdot k}$ and the overdispersion factor $\theta_m$, as well as the rate parameters $\lambda_{\bigcdot u \bigcdot}$ associated with each Markov process. This factorization results in a variational posterior of the form:
\begin{multline}
    q(\Theta,Z) = q(c_0)q(d_0)\prod_m q(\theta_m) \prod_u q(s_u) q(\lambda_{0u}) \prod_r q(\lambda_{xur}) \times \\
    \prod_k q(c_k) q(d_k)
    q(\lambda_{zuk}) q(c_{zk}) q(d_{zk}) q(z_k) q(\pi_k) q(A_k)
\end{multline}
With this ansatz, the variational objective decomposes in a natural way, and choices are available for nearly all of the $q$s that lead to closed-form updates.

\subsection*{Variational posterior}
From Eq (\ref{obs_model}) and (\ref{fr_model}) above, we can write the probability of the observed data $N$ as
\begin{multline}
    \label{log_evidence}
    \log p(N, z|x, \Theta) = \sum_{mkr} \left[
        N_m \left( \log \theta_m +
            \log \lambda_{0u(m)} +
            z_{t(m) k} \log \lambda_{zu(m) k} +
            x_{t(m) r} \log \lambda_{xu(m) r}
            \right)
    \right] \\
    - \sum_m \theta_m \Lambda_{t(m) u(m)} +
    \sum_{mk} \log (A_k)_{z_{t(m)+1, k}, z_{t(m), k}} +
    \sum_k \log (\pi_k)_{z_{0k}} + \text{constant,}
\end{multline}
where again, $m$ indexes observations of $(t(m),u(m))$ pairs and the last two nontrivial terms represent the probability of the Markov sequence given by $z_{tk}$. Given that Eq (\ref{log_evidence}) is of an exponential family form for $\theta$ and $\lambda$ when conditioned on all other variables, free-form variational arguments \cite{Wainwright2008-ii} suggest variational posteriors:
\begin{align}
    \lambda_{0u} &\sim \text{Gamma}(\alpha_{0u}, \beta_{0u}) &
    \lambda_{zuk} &\sim \text{Gamma}(\alpha_{zuk}, \beta_{zuk}) &
    \lambda_{xur} &\sim \text{Gamma}(\alpha_{xur}, \beta_{xur})
\end{align}
For the first of these two, updates in terms of sufficient statistics involving expectations of $\gamma = (c, d)$ are straightforward (see Supplement). However, this relies on the fact that $z_t \in \lbrace0, 1\rbrace$. The observed covariates $x_t$ follow no such restriction, which results in a transcendental equation for the $\beta_x$ updates. In our implementation of the model, we solve this using an explicit BFGS optimization on each iteration. Moreover, we place non-hierarchical Gamma priors on these effects: $\lambda_{xur} \sim \text{Gamma}(a_{xur}, b_{xur})$.

As stated above, for the latent states and baselines, we assume hierarchical priors. This allows us to model each neuron's firing rate response to a particular stimulus as being drawn from a population response to that same stimulus. We also assume that the moment-to-moment noise in firing rates, $\theta_m$, follows a neuron-specific distribution. As a result of the form of this hierarchy given in Eq (\ref{hierarchy}), the first piece in Eq (\ref{elbo}) contains multiple terms of the form
\begin{equation}
    \mathbb{E}_q \left[\sum_u \log p(\lambda_u|c, d)\right] = \sum_u \mathbb{E}_q \left[
    (c - 1) \log \lambda_u - cd\lambda_u + c \log cd - \log \Gamma(c)
    \right]
\end{equation}
In order to calculate the expectation, we make use of the following inequality \cite{abramowitz1964handbook}
\begin{equation}
    \sqrt{2\pi} \le \frac{z!}{z^{z+\frac{1}{2}} e^{-z}} \le e
\end{equation}
to lower bound the negative gamma function and approximate the above as
\begin{equation}
    \log p(\lambda) \ge \sum_u \left[
    (c - 1) (\log \lambda_u + 1) - cd\lambda_u + c \log d + \frac{1}{2}\log c\right]
\end{equation}
Clearly, the conditional probabilities for $c$ and $d$ are gamma in form, so that if we use priors $c \sim \text{Gamma}(a_c, b_c)$ and $d\sim \text{Gamma}(a_d, b_d)$ the posteriors have the form
\begin{align}
    c &\sim \text{Gamma}\left(a_c + \frac{U}{2},
    b_c + \sum_u\mathbb{E}_q
        \left[d \lambda_u - \log \lambda_u - \log d - 1\right]\right) \\
    d &\sim \text{Gamma}\left(
        a_d + U\mathbb{E}_q[c], b_d + \sum_u \mathbb{E}_q [c \lambda_u]
    \right)
\end{align}
This basic form, with appropriate indices added, gives the update rules for the hyperparameter posteriors for $\lambda_0$ and $\lambda_z$. For $\theta$, we simply set $c = s_u$ and $d = 1$.

For each latent variable $z$, the Markov Chain parameters $\pi_k$ and $A_k$, together with the observation model Eq (\ref{log_evidence}) determine a Hidden Markov Model, for which inference can be performed efficiently via conjugate updates and the well-known forward-backward algorithm \cite{beal2003variational}. More explicitly, given $\pi$, $A$, and the emission probabilities for the observations, $\log p(N|z)$, the forward-backward algorithm returns the probabilities $p(z_t=s)$ (posterior marginal), $p(z_{t+1} =s', z_t=s)$ (two-slice marginal) and $\log Z$ (normalizing constant).

\section*{Experiments}
\subsection*{Synthetic data}
We generated synthetic data from the model for $U=100$ neurons for $T=10,000$ time bins of $dt=0.0333s$ ($\approx 6$min of movie at 30 frames per second). Assumed firing rates and effect sizes were realistic for cortical neurons, with mean baseline rates of 10 spikes/s and firing rate effects given by a $\text{Gamma}(1, 1)$ distribution for $K_{\text{data}}=3$ latent features. In addition, we included $R=3$ known covariates generated according to Markov dynamics. For this experiment, we assumed that each unit was presented only once with the stimulus time series, so that $M_{tu} = 1$. That is, we tested a case in which inference was driven primarily by variability in population responses across stimuli rather than pooling of data across repetitions of the same stimulus. Moreover, to test the model's ability to parsimoniously infer features, we set $K=5$. That is, we asked the model to recover more features than were present in the data. Finally, we placed hierarchical priors on neurons' baseline firing rates and sparse hierarchical priors on firing rate effects of latent states. We used 10 random restarts and iterated over parameter updates until the fractional change in $\mathcal{L}$ dropped below $10^{-4}$.

\begin{figure}[!ht]
    \includegraphics[width=\linewidth]{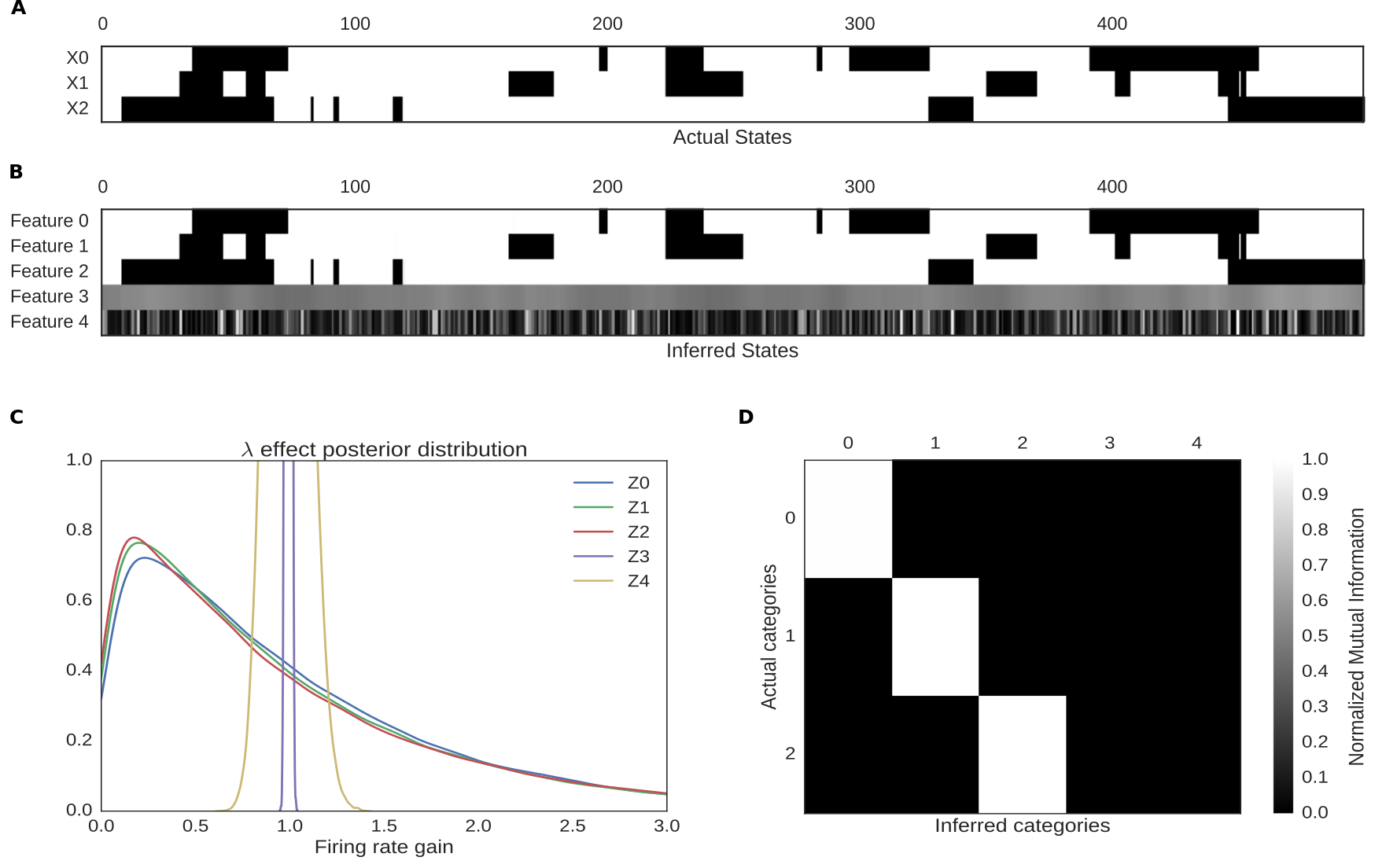}
	\caption{\bf Comparison of actual and inferred states of the synthetic data.} A: Actual features for a subset of stimulus times in the synthetic dataset. B: Recovered binary features for the same subset. Features have been reordered for display. The unused features are in gray, indicating a high posterior uncertainty in the model. C: Population posterior distributions for inferred hyper parameters. Features 3 and 4 are effectively point masses around gain 1 (no effect), while features 1--3 approximate the $\text{Gamma}(1, 1)$ data-generating model. D: Normalized mutual information between actual and inferred states.
	\label{synthetic}
\end{figure}

As seen in Figure \ref{synthetic}, the model correctly recovers only the features present in the original data. We quantified this by calculating the normalized mutual information $\hat{I}\equiv I(X, Y)/\sqrt{H(X)H(Y)}$, between the actual states and the inferred states, with $H(Z)$ and $I$ estimated by averaging the variational posteriors (both absolute and conditioned on observed states) across time. Note that superfluous features in the model have high posterior uncertainty for $z_k$ and high posterior confidence for $\lambda_{zk}$ around 1 (no effect). In addition, the model correctly recovers coefficients for the observed covariates, and when limited to fewer features than in the generating model, recovers a subset of the features accurately rather than blending features together (Figure \ref{synthetic}).

\subsection*{Labeled neural data}
\begin{figure}[!ht]
    \includegraphics[width=\linewidth]{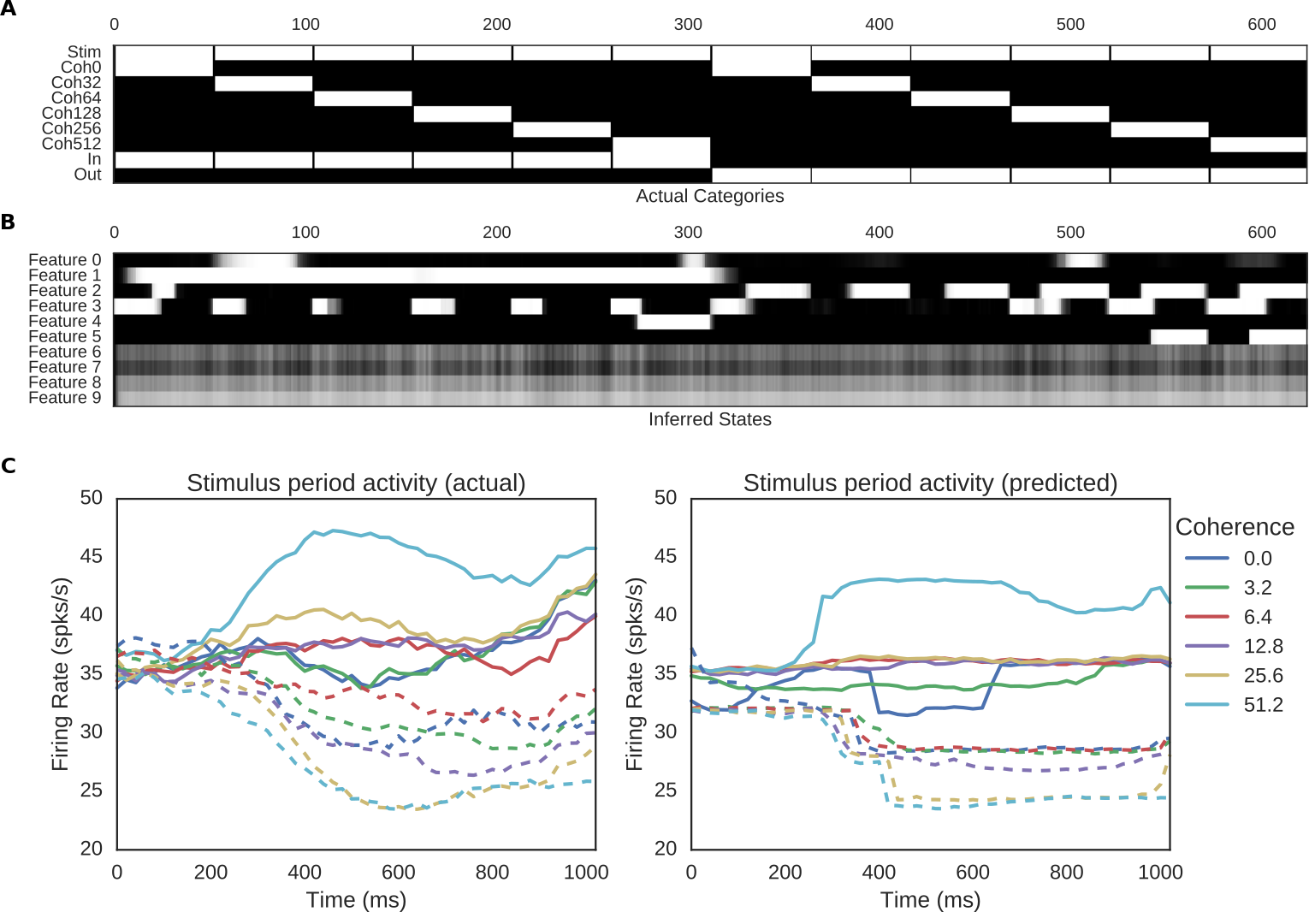}
	\caption{\bf Comparison of actual and inferred states of the Roitman dataset.}
	A: Indicators of actual categories represented in the stimulus presentation period. The categories are not independent of each other. Stimuli are joined sequentially, labeled by a single, unique stimulus time. B: Recovered binary features during the stimulus presentation period. Note that model features 6 -- 9 are unused and that Features 1 \& 2 closely track the In and Out features of the data, respectively. C: Actual and predicted firing rates for the stimulus period. Note that the model infers stimulus categories from the data, including appropriate timing of differentiation between categories.
	\label{roitman}
\end{figure}

We applied our model to a well-studied neural data set comprising single neuron recordings from macaque area LIP collected during the performance of a perceptual discrimination task \cite{roitman2002response}\footnote{Data available at \texttt{https://www.shadlenlab.columbia.edu/resources/RoitmanDataCode.html}}. In the experiment, stimuli consisted of randomly moving dots, some percentage of which moved coherently in either the preferred or anti-preferred direction of motion for each neuron. The animal's task was to report the direction of motion. Thus, in addition to 5 coherence levels, each trial also varied based on whether the motion direction corresponded to the target in or out of the response field as depicted in Fig. \ref{roitman}.\footnote{In the case of 0\% coherence, the direction of motion was inherently ambiguous and coded according to the monkey's eventual choice.}

We fit a model with $K = 10$ features and $U = 27$ units to neural responses from the 1-second stimulus presentation period of the task. Spike counts corresponded to bins of $dt = 20$ms. For this experiment, units were individually recorded, so each unit experienced a different number of presentations of each stimulus condition, implying a ragged observation matrix. As a result, this dataset tests the model's ability to leverage shared task structure across multiple sessions of recording, demonstrating that simultaneously recorded units are not required for inference of latent states.

Figure \ref{roitman} shows the experimental labels from the concatenated stimulus periods, along with labels inferred by our model. Once again, the model has left some features unused, but correctly discerned differences between stimuli in the unlabeled data. Even more importantly, though given the opportunity to infer ten distinct stimulus classes, the model has made use of only five. Moreover, the discovered features clearly recapitulate the factorial design of the experiment, with the two most prominent features, $Z_1$ and $Z_2$, capturing complementary values of the variable with the largest effect in the experiment: whether or not the relevant target was inside our outside the receptive field of the recorded neuron. This difference can be observed in both the averaged experimental data and the predicted data from the model (see Figure \ref{roitman}.C), where the largest differences are between the dotted and solid lines.

But the model also reproduces less obvious features: it correctly discriminates between two identical stimulus conditions (0\% coherence) based on the monkey's eventual decision (In vs Out). In addition, the model correctly captures the initial 200ms ``dead time'' during the stimulus period, in which firing rates remain at pre-stimulus baseline. (Note that the timing is locked to the stimulus and consistent across trials, not idiosyncratic to each trial as in \cite{Latimer2015-pb}.) Finally, the model resists detection of features with little support in the experimental data. For instance, while feature $Z_4$ captures the large difference between 50\% coherence and other stimuli, the model does not infer a difference between intermediate coherence levels that are indistinguishable in this particular dataset. That is, mismatches between ground truth labels and model-inferred features here reflect underlying ambiguities in the neural data, while the model's inferred features correctly pick out those combinations of variables most responsible for differences in spiking across conditions.

\subsection*{Visual category data}
\label{it_neuron_expt}
\begin{figure}[!ht]
    \includegraphics[width=\linewidth]{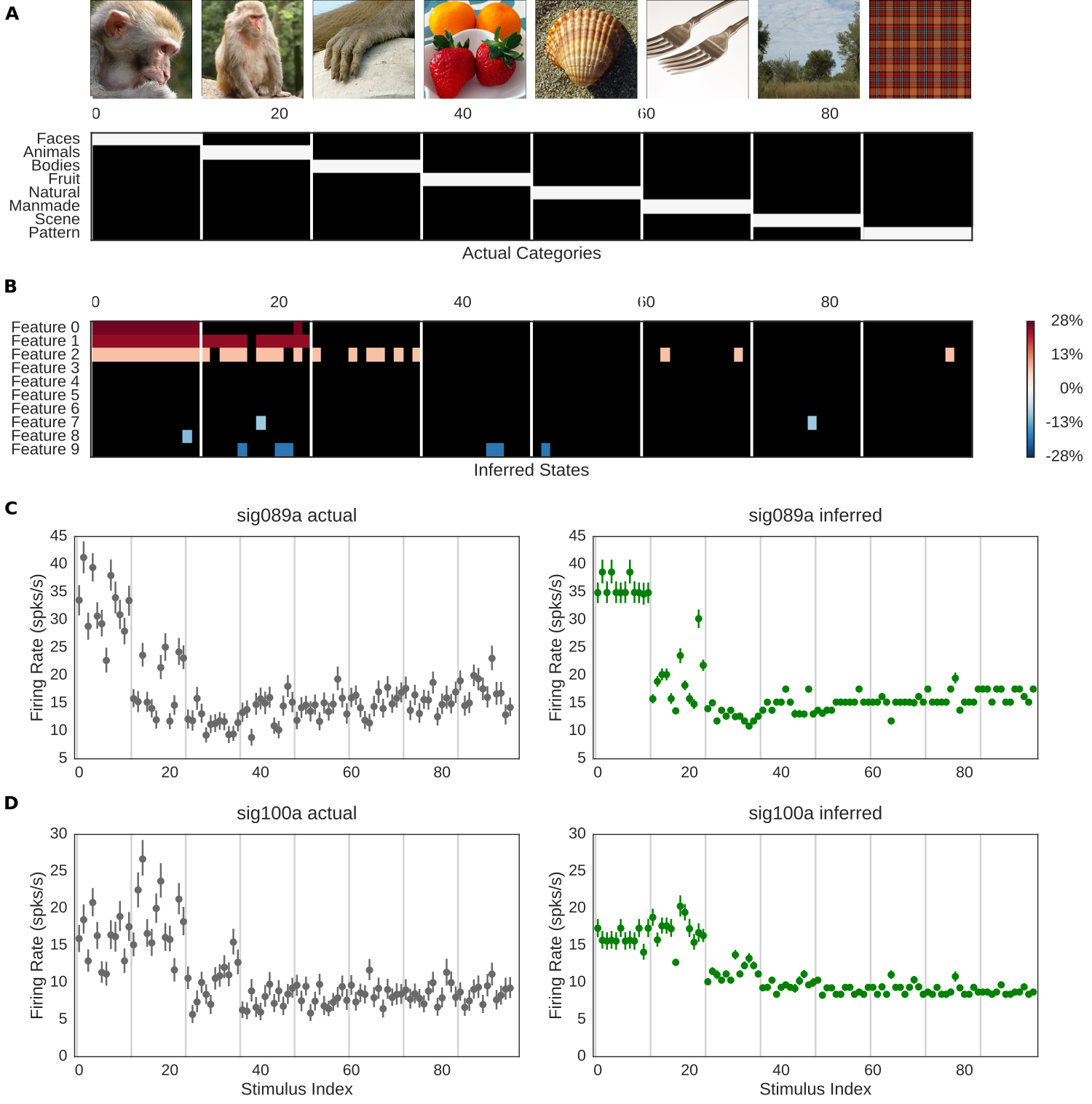}
	\caption{\bf Comparison of actual and inferred states of the macaque dataset.}
	A: Actual categories in the macaque data set. 96 stimuli comprising 8 categories were presented in 1483 trials, with each stimulus presented to each neuron $\sim$15 times. B: The inferred states from our model. The color represents the multiplicative effect to the population baseline firing rates. Note that our model shows a clear increase of firing rates in the categories of monkey faces, whole monkeys, and some stimuli within monkey bodies. C: Actual and predicted spikes per second across all stimulus of neuron 089a. D. Actual and predicted spikes per second across all stimulus of neuron 100a. Error bars for data represent 95\% credible intervals for the firing rate based on observed data under a Poisson model with weak priors. Error bars on predictions are 95\% credible intervals based on simulation from the approximate posterior for the given unit.
	\label{fig:imgclust}
\end{figure}

As a second test of our model, we applied our algorithm to a designed structured stimuli dataset comprising $U = 56$ neurons from macaque inferotemporal cortex \cite{McMahon2014-qq}. These neurons were repeatedly presented with 96 stimuli comprising 8 categories ($M$ = 1483 total trials, with each stimulus exposed between 12 to 19 times to each unit) comprising monkey faces, monkey bodies, whole monkeys, natural scenes, food, manmade objects, and patterns (Figure \ref{fig:imgclust}.A). Data consisted of spike time series, which we binned into a 300ms pre-stimulus baseline, a 300ms stimulus presentation period, and a 300ms post-stimulus period. Three trials were excluded because of the abnormal stimulus presentation period. To maximize interpretability of the results, we placed strong priors on the $\pi_k$ to formalize the assumption that all features were off during the baseline period. We also modeled overdispersion with extremely weak priors to encourage the model to attribute fluctuations in firing to noise in preference to feature detection. We again fit $K = 10$ features with sparse hierarchical priors on population responses.

\begin{figure}[!ht]
    \includegraphics[width=\linewidth]{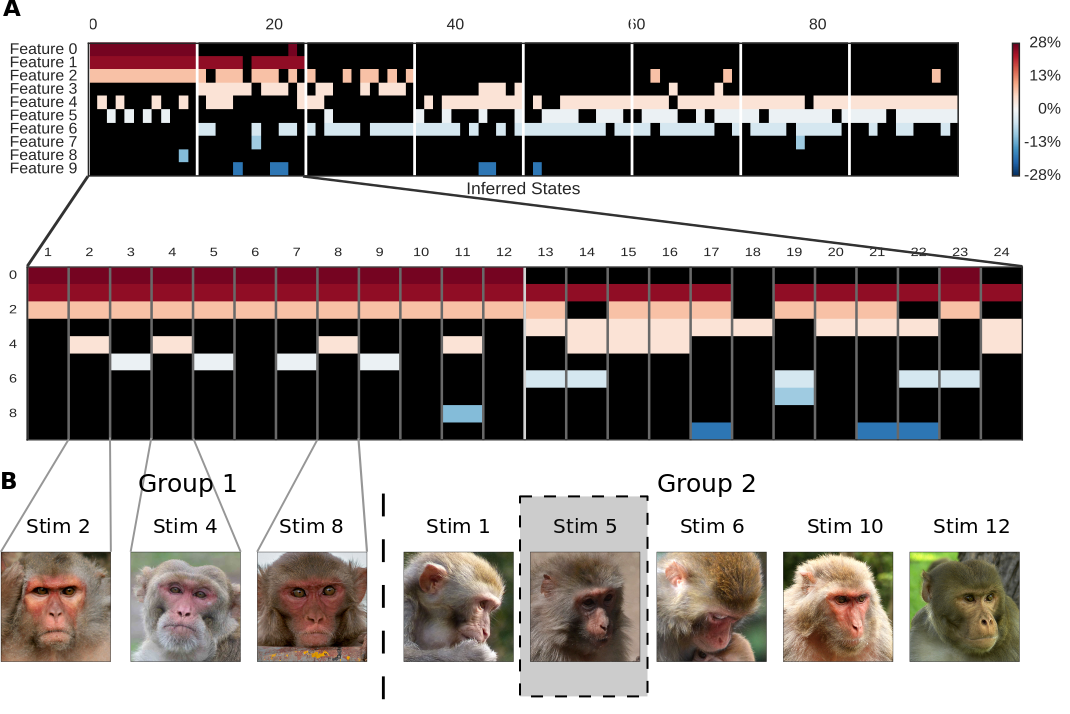}
	\caption{\bf Small features suggest additional neural hypotheses.}
	A: Zoomed-in view of Figure \ref{fig:imgclust}.A, focusing on the first 24 images. B: The feature combinations 0\&1\&2 (Group 1) and 0\&1\&2\&4 (Group 2) are distinguished by direct vs. indirect gaze. Stimulus 5, which is coded 0\&1\&2\&5, may be considered a false negative.
	\label{fig:imgclust_sub}
\end{figure}

The inferred categories based on binned population responses are shown in Figure \ref{fig:imgclust}.B. For clarity, in Figure \ref{fig:imgclust}, we only show population mean effects with a ``$>5\%$ gain'' modulation sorted from the highest to the lowest, though the full set of inferred states can be found in Figure \ref{fig:imgclust_sub}. Out of the original categories, our model successfully recovers three features clearly corresponding to categories involving monkeys (Features 0 -- 2). These can be viewed additively, with Feature 0 exclusive to monkey face close-ups, Feature 1 any photo containing a monkey face, either near or far; and Feature 2 any image containing a monkey body part (including faces); but as we will argue, given the nature of the model, it may be better to view these as a ``combinatorial'' code, with monkey close-ups encoded as $0\& 1\& 2$ ($\sim 59.46\%$ increase in firing), whole monkeys as $1\& 2$ ($\sim 32.47\%$ increase), and monkey body parts as $2$ ($\sim 7.62\%$ increase). Of course, this is consistent with what was found in \cite{McMahon2014-qq}, though our model used no labels on the images. And our interpretation that these neurons are sensitive to close-ups and faraway face and body parts is consistent with findings by another study using different experimental settings \cite{McMahon5537}.

Again, as noted above, our results in Figure \ref{fig:imgclust}.A and \ref{fig:imgclust}.B indicate predicted population responses, derived from the hierarchical prior. As evidenced in Figure \ref{fig:imgclust}.C and \ref{fig:imgclust}.D, individual neuron effects could be much larger. These panels show data for two example units, along with the model's prediction. Clearly, the model recapitulates the largest distinctions between images in the data, though the assumption that firing rates should be the same for all images with similar features fails to capture some variability in the results. Even so, uncertainties in the predicted firing rates are also in line with uncertainties from those of observed rates, indicating that our model is correctly accounting for trial-to-trial noise.

Finally, even the weaker, sparser features inferred by our model captured intriguing additional information. As shown in Figure \ref{fig:imgclust_sub}, Feature 4, a feature only weakly present in the population as a whole (and thus ignored in \ref{fig:imgclust_sub}.A), when combined with the stronger Features 0, 1, and 2, successfully distinguishes between the monkey close-ups with direct and averted gaze. (Stimulus 5, with averted gaze, is additionally tagged with Feature 5, which we view as an imperfect match.) Thus, despite the fact that Feature 4 is barely a 3.4\% gain change over the population, it suggests a link between neural firing and gaze direction, one for which there happens to be ample evidence \cite{Perrett23, Freiwald845}. Similarly, Feature 5, barely a 1.1\% effect, correctly tags three of the four close-ups with rightward gaze (with one false positive). Clearly, neither of these results is dispositive in this particular dataset, but in the absence of hypotheses about the effect of head orientation and gaze on neuronal firing, these minor features might suggest hypotheses for future experiments.

An additional feature of our approach is that the generated labels provide a concise and fairly complete summary of the stimulus-related activity of all neural recordings, which can be observed by comparing the categorization performance of decoded neural activity to the categorization performance of the decoded features. Although our model is not a data compression method, it nonetheless preserves most of the information about image category contained in the $N=56$ dimensional spike counts via a 10-dimensional binary code. That is, using a sparse logistic regression on two-bit and three-bit combinations of our features to predict stimulus category ties and outperforms, respectively a multinomial logistic regression on the raw spike counts (see Supplement).

\section*{Discussion}

Here, we have proposed and implemented a method for learning features in stimuli via the responses of populations of spiking neurons. This work addresses a growing trend in systems neuroscience --- the increasing use of rich and unstructured or structured stimulus sets --- without requiring either expert labeling or a metric on the stimulus space. As such, we expect it to be of particular use in disciplines like social neuroscience, olfaction, and other areas in which the real world is complex and strong hypotheses about the forms of the neural code are lacking. By learning features of interest to neural populations directly from neural data, we stand to generate unexpected, more accurate (less biased) hypotheses regarding the neural representation of the external world.

Here, we have validated this method using structured, labeled stimuli more typical of neuroscience experiments, showing that our model is capable of parsimoniously and correctly inferring features in the low signal-to-noise regime of cortical activity, even in the case of independently recorded neurons. Furthermore, by employing a fully variational, Bayesian approach to inference, we gain three key advantages: First, we gain the advantages of Bayesianism in general: estimates of confidence in inferences, parsimony and regularization via priors, and the ability to do principled model comparison. Second, variational methods scale well to large datasets and can be easily parallelized when combining data from multiple recording sessions. Finally, variational methods are fast, in that they typically converge within only a few tens of iterations and in many case (such as ours) require mostly simple coordinate updates.

Finally, even small features in our model recapitulated known physiological results regarding face encoding in single neurons. And while these features alone might not provide proof positive of, e.g., viewpoint tuning, similar findings would be valuable in generating hypotheses in cases where the stimulus space and its neural correlates remain poorly understood. Thus our model facilitates an iterative experimental process: subjects are first be exposed to large, heterogeneous data; stimuli are then tagged based on neural responses; and finally, features with the largest effects are used to refine the set until it most accurately represents those stimuli with the largest neural correlates. Combined with the modularity of this and similar approaches, such models provide a promising opportunity to ``build out'' additional features that will meet the challenges of the next generation of experimental data.

\section*{Acknowledgments}
We would like to thank David McMahon and David Leopold for generously sharing the visual cateogry stimuli and neural data from \cite{McMahon2014-qq} and for comments on the manuscript.

\bibliography{chen_beck_pearson}{}

\begin{thebibliography}{10}

\bibitem{steveninck1988realtime}
Steveninck RDRV, Bialek W.
\newblock Real-Time Performance of a Movement-Sensitive Neuron in the Blowfly
  Visual System: Coding and Information Transfer in Short Spike Sequences.
\newblock Proceedings of the Royal Society of London B: Biological Sciences.
  1988;234(1277):379--414.
\newblock doi:{10.1098/rspb.1988.0055}.

\bibitem{ringach2004reverse}
Ringach D, Shapley R.
\newblock Reverse correlation in neurophysiology.
\newblock Cognitive Science. 2004;28(2):147--166.

\bibitem{ringach2002receptive}
Ringach DL, Hawken MJ, Shapley R.
\newblock Receptive field structure of neurons in monkey primary visual cortex
  revealed by stimulation with natural image sequences.
\newblock Journal of vision. 2002;2(1):2.

\bibitem{sharpee2004analyzing}
Sharpee T, Rust NC, Bialek W.
\newblock Analyzing neural responses to natural signals: maximally informative
  dimensions.
\newblock Neural computation. 2004;16(2):223--250.

\bibitem{Vinje2000-dx}
Vinje WE, Gallant JL.
\newblock Sparse coding and decorrelation in primary visual cortex during
  natural vision.
\newblock Science. 2000;287(5456):1273--1276.

\bibitem{Williamson2013-rg}
Williamson RS, Sahani M, Pillow JW.
\newblock The equivalence of information-theoretic and likelihood-based methods
  for neural dimensionality reduction. 2013;.

\bibitem{Vu2011-da}
Vu VQ, Ravikumar P, Naselaris T, Kay KN, Gallant JL, Yu B.
\newblock Encoding and Decoding V1 fMRI Responses to Natural Images with Sparse
  Nonparametric Models.
\newblock Ann Appl Stat. 2011;5(2B):1159--1182.

\bibitem{Huth2012-cj}
Huth AG, Nishimoto S, Vu AT, Gallant JL.
\newblock A continuous semantic space describes the representation of thousands
  of object and action categories across the human brain.
\newblock Neuron. 2012;76(6):1210--1224.

\bibitem{Stansbury2013-nm}
Stansbury DE, Naselaris T, Gallant JL.
\newblock Natural scene statistics account for the representation of scene
  categories in human visual cortex.
\newblock Neuron. 2013;79(5):1025--1034.

\bibitem{Pillow2008-em}
Pillow JW, Shlens J, Paninski L, Sher A, Litke AM, Chichilnisky EJ, et~al.
\newblock Spatio-temporal correlations and visual signalling in a complete
  neuronal population.
\newblock Nature. 2008;454(7207):995--999.

\bibitem{Vogelstein2009-ax}
Vogelstein JT, Watson BO, Packer AM, Yuste R, Jedynak B, Paninski L.
\newblock Spike inference from calcium imaging using sequential Monte Carlo
  methods.
\newblock Biophys J. 2009;97(2):636--655.

\bibitem{Park2014-el}
Park IM, Meister MLR, Huk AC, Pillow JW.
\newblock Encoding and decoding in parietal cortex during sensorimotor
  decision-making.
\newblock Nat Neurosci. 2014;17(10):1395--1403.

\bibitem{Buesing2014-ta}
Buesing L, Machado TA, Cunningham JP, Paninski L.
\newblock Clustered factor analysis of multineuronal spike data.
\newblock In: Ghahramani Z, Welling M, Cortes C, Lawrence ND, Weinberger KQ,
  editors. Advances in Neural Information Processing Systems 27. Curran
  Associates, Inc.; 2014. p. 3500--3508.

\bibitem{Archer2015-ec}
Archer E, Park IM, Buesing L, Cunningham J, Paninski L.
\newblock Black box variational inference for state space models. 2015;.

\bibitem{Zhao2016-bw}
Zhao Y, Park IM.
\newblock Variational Latent Gaussian Process for Recovering {Single-Trial}
  Dynamics from Population Spike Trains. 2016;.

\bibitem{Gao2016-ck}
Gao Y, Archer E, Paninski L, Cunningham JP.
\newblock Linear dynamical neural population models through nonlinear
  embeddings. 2016;.

\bibitem{Ulrich2014-zc}
Ulrich KR, Carlson DE, Lian W, Borg JS, Dzirasa K, Carin L.
\newblock Analysis of Brain States from {Multi-Region} {LFP} {Time-Series}.
\newblock In: Ghahramani Z, Welling M, Cortes C, Lawrence ND, Weinberger KQ,
  editors. Advances in Neural Information Processing Systems 27. Curran
  Associates, Inc.; 2014. p. 2483--2491.

\bibitem{Putzky2014-up}
Putzky P, Franzen F, Bassetto G, Macke JH.
\newblock A Bayesian model for identifying hierarchically organised states in
  neural population activity.
\newblock In: Ghahramani Z, Welling M, Cortes C, Lawrence ND, Weinberger KQ,
  editors. Advances in Neural Information Processing Systems 27. Curran
  Associates, Inc.; 2014. p. 3095--3103.

\bibitem{Wainwright2008-ii}
Wainwright MJ, Jordan MI.
\newblock Graphical Models, Exponential Families, and Variational Inference.
\newblock Found Trends Mach Learn. 2008;1(1-2):1--305.

\bibitem{ghahramani1997factorial}
Ghahramani Z, Jordan MI.
\newblock Factorial hidden Markov models.
\newblock Machine learning. 1997;29(2-3):245--273.

\bibitem{abramowitz1964handbook}
Abramowitz M, Stegun IA.
\newblock Handbook of mathematical functions: with formulas, graphs, and
  mathematical tables.
\newblock 55. Courier Corporation; 1964.

\bibitem{beal2003variational}
Beal MJ.
\newblock Variational algorithms for approximate Bayesian inference.
\newblock University of London; 2003.

\bibitem{roitman2002response}
Roitman JD, Shadlen MN.
\newblock Response of neurons in the lateral intraparietal area during a
  combined visual discrimination reaction time task.
\newblock The Journal of neuroscience. 2002;22(21):9475--9489.

\bibitem{Latimer2015-pb}
Latimer KW, Yates JL, Meister MLR, Huk AC, Pillow JW.
\newblock Single-trial spike trains in parietal cortex reveal discrete steps
  during decision-making.
\newblock Science. 2015;349(6244):184--187.

\bibitem{McMahon2014-qq}
McMahon DBT, Jones AP, Bondar IV, Leopold DA.
\newblock Face-selective neurons maintain consistent visual responses across
  months.
\newblock Proc Natl Acad Sci U S A. 2014;111(22):8251--8256.

\bibitem{McMahon5537}
McMahon DBT, Russ BE, Elnaiem HD, Kurnikova AI, Leopold DA.
\newblock Single-Unit Activity during Natural Vision: Diversity, Consistency,
  and Spatial Sensitivity among AF Face Patch Neurons.
\newblock Journal of Neuroscience. 2015;35(14):5537--5548.
\newblock doi:{10.1523/JNEUROSCI.3825-14.2015}.

\bibitem{Perrett23}
Perrett DI, Hietanen JK, Oram MW, Benson PJ, Rolls ET.
\newblock Organization and Functions of Cells Responsive to Faces in the
  Temporal Cortex [and Discussion].
\newblock Philosophical Transactions of the Royal Society of London B:
  Biological Sciences. 1992;335(1273):23--30.
\newblock doi:{10.1098/rstb.1992.0003}.

\bibitem{Freiwald845}
Freiwald WA, Tsao DY.
\newblock Functional Compartmentalization and Viewpoint Generalization Within
  the Macaque Face-Processing System.
\newblock Science. 2010;330(6005):845--851.
\newblock doi:{10.1126/science.1194908}.

\bibitem{Yu2006-bb}
Yu SZ, Kobayashi H.
\newblock Practical implementation of an efficient forward-backward algorithm
  for an explicit-duration hidden Markov model.
\newblock Signal Processing, IEEE Transactions on. 2006;54(5):1947--1951.

\bibitem{Blei2006-oh}
Blei DM, Jordan MI.
\newblock Variational inference for Dirichlet process mixtures.
\newblock Bayesian Anal. 2006;1(1):121--143.

\bibitem{Mitchell1993-sl}
Mitchell CD, Jamieson LH.
\newblock Modeling Duration in a Hidden Markov Model with the Exponential
  Family.
\newblock In: Acoustics, Speech, and Signal Processing; 1993.

\bibitem{Mitchell1995-go}
Mitchell C, Harper M, Jamieson L.
\newblock On the complexity of explicit duration {HMM's}.
\newblock IEEE Trans Audio Speech Lang Processing. 1995;3(3):213--217.

\end{thebibliography}


\newcommand{\lorem}{{\bf LOREM}}
\newcommand{\ipsum}{{\bf IPSUM}}

\newcommand{\ud}{\mathrm{d}} 
\newcommand{\R}{\mathbb{R}} 
\newcommand{\T}{\mathscr{T}}

\newcommand{\I}{\mathbb{I}} 

\newcommand{\CE}[2]{\mathrm{E}\left[\,#1\,|\,#2\,\right]} 

\newcommand{\by}{\boldsymbol{y}}
\newcommand{\bc}{\boldsymbol{c}}
\newcommand{\bd}{\boldsymbol{d}}
\newcommand{\bP}{\boldsymbol{\Phi}}
\newcommand{\bZ}{Z}
\newcommand{\bV}{V}
\newcommand{\br}{\boldsymbol{r}}
\newcommand{\bt}{\boldsymbol{t}}
\newcommand{\btheta}{\boldsymbol{\vartheta}}
\newcommand{\bht}{\hat{\boldsymbol{\vartheta}}}
\newcommand{\bz}{\boldsymbol{z}}
\newcommand{\bx}{{\boldsymbol{x}}}
\newcommand{\bv}{v}
\newcommand{\bw}{\boldsymbol{w}}
\newcommand{\vv}{\boldsymbol{v}}
\newcommand{\bepsilon}{\boldsymbol{\varepsilon}}


\begin{flushleft}
{\Large
\textbf{Supplementary Information 1 \\ Mathematical derivation for ELBO and Inference \\}
}
%
%
\end{flushleft}

\section{Evidence Lower Bound (ELBO)}
Here we derive the evidence lower bound (ELBO) used as a variational objective by our inference algorithm. That is, we want to calculate
\begin{equation}
    \mathcal{L} \equiv \mathbb{E}_q \left[\log \frac{p(\Theta|N)}{q(\Theta)} \right] = \mathbb{E}_q \left[\log p(\Theta|N) \right] + \mathcal{H}[q(\Theta)]
\end{equation}
From \cite{beal2003variational}, this can be written
\begin{multline}
    \mathcal{L} = \mathbb{E}_{q(\pi)} \left[\log \frac{p(\pi)}{q(\pi)} \right]
    + \mathbb{E}_{q(A)} \left[\log \frac{p(A)}{q(A)} \right]
    + \mathbb{E}_{q}\left[ \log \frac{p(N, z|\lambda, A, \pi)}{q(z)}\right] \\
    + \mathbb{E}_{q(\theta)} \left[\log \frac{p(\theta)}{q(\theta)} \right]
    + \mathbb{E}_{q(\lambda)} \left[\log \frac{p(\lambda)}{q(\lambda)} \right]
    + \mathbb{E}_{q(\gamma)} \left[\log \frac{p(\gamma)}{q(\gamma)} \right]
\end{multline}
For the first two terms, updates are standard and covered in \cite{beal2003variational}. The rest we do piece-by-piece below:

\subsection{Log evidence}
\label{sec:log_evidence}
We would like to calculate $\mathbb{E}_{q}\left[ \log \frac{p(N, z|x, \Theta)}{q(z)}\right]$. To do this, we make use of expectations calculated via the posteriors returned from the forward-backward algorithm
\begin{align}
    \xi_{t} &\equiv p(z_{t}|N, \theta) &
    \Xi_{t, ij} &\equiv p(z_{t+1} = j, z_{t} = i|N, \theta) &
    \log Z_{t} &= \log p(N_{t+1}|N_{t}, \Theta)
\end{align}
Here, we have suppressed the latent feature index $k$ and abuse notation by writing the observation index as $t$, but in the case of multiple observations at a given time, we pool across units and presentations: $N_t \equiv \sum_{m; t(m) = t} N_m$. From this, we can write
\begin{multline}
    \label{eq:log_evidence}
     \mathbb{E}_{q}\left[ \log p(N, z|x, \Theta) \right] =
     \sum_{mkr} \left[N_m \left(
        \overline{\log \theta_m} + \overline{\log \lambda_{0u(m)}} +
        \xi_{t(m)k}\overline{\log \lambda_{zuk}} +
        x_{t(m)r} \overline{\log \lambda_{xu(m)r}}
     \right) \right] \\
     - \sum_m \overline{\theta_m} \mathbb{E}_q\left[\Lambda_{t(m)u(m)} \right]
      + \sum_{tk} \left[ \text{tr}\left(\Xi_{tk} \overline{\log A_k^T} \right)
     + \xi_{0k}^T \overline{\log \pi_k}
     \right]
    + \text{constant}
 \end{multline}
In what follows, we will drop the irrelevant constant. For $\overline{\log y}$, where $y \in \lbrace \theta, \lambda_0, \lambda_z, \lambda_x \rbrace$, the assumption $q(y) = \text{Ga}(\alpha, \beta)$ gives
\begin{equation}
    \overline{\log y} = \psi(\alpha) - \log \beta
\end{equation}
with $\psi(x)$ the digamma function. Likewise, the expectation $\overline{\theta}$ is straightforward. For the expectation of the rate, we have
\begin{equation}
    \label{eff_rate}
    \mathbb{E}_q\left[
        \lambda_{0u} \prod_k (\lambda_{zuk})^{z_{tk}} \prod_r (\lambda_{xur})^{x_{tr}}
    \right] = \frac{\alpha_{0u}}{\beta_{0u}}
    \prod_k \left(1 - \xi_{tk} + \xi_{tk} \frac{\alpha_{zuk}}{\beta_{zuk}} \right)
    \prod_r \frac{1}{\beta_{xur}^{x_{tr}}} \frac{\Gamma(\alpha_{xur} + x_{tr})}{\Gamma(\alpha_{xur})}
\end{equation}
However, for $\alpha \gg x$, we have $\Gamma(\alpha + x)/\Gamma(\alpha) \approx \alpha^x$, so that we can write
\begin{equation}
    \label{HFG}
    \mathbb{E}_q[\Lambda_{tu}] = H_{0u} F_{tu} G_{tu}
\end{equation}
with $G_{tu} \approx \prod_r (\alpha_{xur}/\beta_{xur})^{x_{tr}}$. In addition, it will later be useful to have the expectation over \emph{all except} a particular feature $k$ or $r$, for which we define
\begin{align}
    \label{F}
    F_{tuk} &\equiv \prod_{k'\neq k} \left(1 - \xi_{tk'} + \xi_{tk'} \frac{\alpha_{zuk'}}{\beta_{zuk'}} \right) \\
    \label{G}
    G_{tur} &\equiv \prod_{r' \neq r} \left(\frac{\alpha_{xur'}}{\beta_{xur'}} \right)^{x_{tr'}}
\end{align}
Finally, we want the entropy of the variational posterior over $z$, $\mathbb{E}_q[-\log q(z)]$. We can write this in the form
\begin{equation}
    -\sum_{tk} \left[
        \xi_{tk}^T\eta_{tk} + \text{tr}\left(\Xi_{tk} \tilde{A}_k^T \right)
        + \xi_{0k}^T\tilde{\pi}_k
        - \log Z_{tk}
    \right]
\end{equation}
with $(\eta, \tilde{A}, \tilde{\pi})$ the parameters of the variational posterior corresponding to the emission, transition, and initial state probabilities of the Markov chain (interpreted as matrices) and $Z$ the normalization constant. From \cite{beal2003variational}, we have that variational updates should give
\begin{align}
    \tilde{A}_k &\leftarrow \overline{\log A_k} &
    \tilde{\pi}_k &\leftarrow \overline{\log \pi_k}
\end{align}
while the effective emission probabilities in the ``on" ($z = 1$) state of the HMM are
\begin{align}
    \eta_{tk} &\leftarrow \delta_{z_{tk}, 1} \sum_{m; t(m) = t} N_m \overline{\log \lambda_{zu(m)k}}
    - \sum_{m; t(m) = t} \overline{\theta_m} H_{0u(m)} F_{tku(m)} G_{tu(m)}
\end{align}
Given these update rules, we can then alternate between calculating $(\eta, \tilde{A}, \tilde{\pi})$, performing forward-backward to get $(\xi, \Xi, \log Z)$ and recalculating $(\eta, \tilde{A}, \tilde{\pi})$.

\subsection{Overdispersion, firing rate effects}
Both the case of $p(\theta)$ and $p(\lambda)$ are straightforward. If we ignore subscripts and write $p(y) = \text{Ga}(a, b)$, $q(y) = \text{Ga}(\alpha, \beta)$, then
\begin{equation}
    \mathbb{E}_q \left[\log \frac{p(y)}{q(y)} \right] =
    (\overline{a} - 1) \overline{\log y} + \overline{b} \overline{y} + \mathcal{H}[q(y)]
\end{equation}
where again, $\overline{\log y}$, $\overline{y}$ and $\mathcal{H}[q(y)]$ are straightforward properties of the Gamma distribution. Expectations of the prior parameters are listed in Table \ref{expectation_table}. Note in the last line that there is no expectation, since we have not assumed a hierarchy over firing rate effects for the covariates, $x$.

\begin{table}[ht]
\caption{Expectations of prior parameters for overdispersion and firing rates}
\label{expectation_table}
\begin{center}
\begin{tabular}{lcc}
\multicolumn{1}{c}{\bf Variable}  &\multicolumn{1}{c}{\bf $\mathbb{E}_q[a]$} &\multicolumn{1}{c}{\bf $\mathbb{E}_q[b]$}
\\ \hline
$\theta$ &$\overline{s}$ &$\overline{s}$ \\
$\lambda_0$ &$\overline{c_0}$ &$\overline{c_0 d_0}$ \\
$\lambda_z$ &$\overline{c_z}$ &$\overline{c_z d_z}$ \\
$\lambda_x$ &$a_x$ &$b_x$ \\
\end{tabular}
\end{center}
\end{table}

\subsection{Hyperparameters}
As shown in the main text, the hyperparameters $c$ and $d$, given gamma priors, have conjugate gamma posteriors, so that their contribution to the evidence lower bound, $\mathbb{E}_q \left[\log \frac{p(\gamma)}{q(\gamma)} \right]$ is a sum of terms of the form
\begin{align}
    (a_{c} - 1) \overline{\log c} + b_{c} \overline{c} + \mathcal{H}[q(c)] +
    (a_{d} - 1) \overline{\log d} + b_{d} \overline{d} + \mathcal{H}[q(d)]
\end{align}
In other words, these are straightforward gamma expectations, functions of the prior parameters $a$ and $b$ for each variable and the corresponding posterior parameters $\alpha$ and $\beta$. Similarly, the overdispersion terms are exactly the same with the substitutions $c \rightarrow s$, $d\rightarrow 1$.

As we will see below, the expectations under the variational posterior of $s$, $c$, and $d$ are themselves straightforward to calculate.

\subsection{Autocorrelated noise}
In the main text, we assumed the $\theta_m$ to be uncorrelated. However, it is possible to model temporal autocorrelation among the $\theta_m$ when observations correspond to the same neuron at successive time points. More specifically, let us replace the observation index $m$ by $\tau$, the experimental clock time. We then write a particular observation as corresponding to a stimulus time $t(\tau)$ and a set of units $u(\tau)$, which, for simplicity, we will assume fixed and simply write as $u$.\footnote{The generalization to partially overlapping neurons and stimuli is straightforward but complex and notationally cumbersome.} To model the autocorrelation of noise across successive times, we then write
\begin{align}
    N_{m} &\sim \text{Pois}(\Lambda_{t(\tau), u} \theta_{\tau u}) &
    \text{Assuming } \theta_{\tau u} = \phi_{\tau u} \theta_{\tau - 1, u}
\end{align}

If $\phi_{0u} = \theta_{0u}$, we then have $\theta_{\tau u} = \prod_{\tau' \le \tau} \phi_{\tau' u}$. In essence, this is a log-autoregressive process in which the innovations are not necessarily normally distributed. In fact, if we further assume that $p(\phi_{\tau u}) = \mathrm{Ga}(s_u, r_u)$ and $q(\phi_{\tau u}) = \mathrm{Ga}(\omega_{\tau u}, \zeta_{\tau u})$, we can once again make use of conjugate updates.

Given these assumptions, we need to make the following modifications for the third and fifth terms in the evidence lower bound of Eq (\ref{eq:log_evidence}):
\begin{align}
    \sum_{mkr} N_m \xi_{t(m)k}\overline{\log \lambda_{zuk}} &\rightarrow
        \sum_{mkr} N_m \xi_{t(m)k}\overline{\log \lambda_{zuk}} +
        \sum_{\tau u} \overline{\log\phi_{\tau u}} \sum_{\tau' \ge \tau} N_{\tau' u} \\
    - \sum_m \overline{\theta_m} \mathbb{E}_q\left[\Lambda_{t(m)u(m)} \right] &\rightarrow
        - \sum_{\tau u} H_{0u} F_{\tau ku} G_{\tau u} \prod_{\tau' \le \tau} \frac{\omega_{\tau' u}}{\zeta_{\tau' u}}
\end{align}

Thus the effective emission probabilities in State 1 now become:
\begin{align}
    \eta_{tk} &\rightarrow \delta_{z_{tk}, 1} \sum_{m} N_m \overline{\log \lambda_{zuk}}
    - \sum_{u; t(\tau) = t} H_{0u} F_{\tau ku} G_{\tau u} \prod_{\tau' \le \tau} \frac{\omega_{\tau' u}}{\zeta_{\tau' u}}
\end{align}

\subsection{Latent states, semi-Markov dynamics}

We model each of the latent states $z_{tk}$ as an independent Markov process for each feature $k$. That is, each $k$ indexes an independent Markov chain with initial state probability $\pi_k\sim \text{Dir}(\alpha_\pi)$ and transition matrix $A_k\sim \text{Dir}(\alpha_A)$. For the semi-Markov case, we assume that the dwell times in each state are distributed independently for each chain according to an integer-valued, truncated lognormal distribution with support on the integers $1\dots D$:
\begin{align}
    \label{semi-markov}
    p_k(d|z = j) &= \text{Log-Normal}(d|m_{jk}, s^2_{jk}) / W_{jk}  \\
    W_{jk} &= \sum_{d = 1}^D \text{Log-Normal}(d|m_{jk}, s^2_{jk})
\end{align}
Note that we have allowed the dwell time distribution to depend on both the feature $k$ and the state of the Markov chain $j$. In addition, we put independent Normal-Gamma priors on the mean $(m_{kj})$ and precision $(\tau_{kj} \equiv s_{kj}^{-2})$ parameters of the distribution: $(m, \tau) \sim \text{NG}(\mu, \lambda, \alpha, \beta)$.

In this case, we additionally need to perform inference on the parameters $(m, \tau)$ of the dwell time distributions for each state. In the case of continuous dwell times, our model in Equation \ref{semi-markov} would have $W = 1$ and be conjugate to the Normal-Gamma prior on $(m, \tau)$, but the restriction to discrete dwell times requires us to again lower bound the variational objective:
\begin{equation}
    \mathbb{E}_q\left[-\log W_{jk} \right] =
    \mathbb{E}_q\left[- \log \left( \sum_{d=1}^D p(d|j)\right) \right]
    \ge -\log \sum_{d = 1}^D \mathbb{E}_q\left[p(d|j)\right]
\end{equation}
This correction for truncation must then be added to $\mathbb{E}_q[p(z|\Theta)]$. For inference in the semi-Markov case, we use an extension of the forward-backward algorithm\cite{Yu2006-bb}, at the expense of computational complexity $\mathcal{O}(SDT)$ $(S = 2)$ per latent state, to calculate $q(z_k)$. For the $4SK$ hyperparameters of the Normal-Gamma distribution, we perform an explicit BFGS optimization on the $4S$ parameters of each chain on each iteration (detailed in Subsection \ref{semimarkovdd}).

\section{Inference}
\subsection{Conjugate updates}
For updates on the overdispersion, firing rate, and hyperparameter variables, we have the simple conjugate update rules depicted in Table \ref{conj_updates}. These can be derived either from free-form variational arguments, or exponential family rules, but are in any case straightforward\cite{Blei2006-oh}. If we assume a $\text{Ga}(a, b)$ prior and $\text{Ga}(\alpha, \beta)$ variational posterior, all of these have a simple algebraic update in terms of sufficient statistics.

\begin{table}[ht]
\caption{Conjugate updates for Gamma distributions}
\label{conj_updates}
\begin{center}
\begin{tabular}{lcc}
\multicolumn{1}{c}{\bf Variable}  &\multicolumn{1}{c}{\bf $\alpha - a$} &\multicolumn{1}{c}{\bf $\beta - b$}
\\ \hline
$\theta_m$         &$N_m$  &$H_{0u(m)}F_{t(m)u(m)}G_{t(m)u(m)}$ \\
$s_u$         &$\frac{1}{2}\sum_m \delta_{u(m), u}$  &$\sum_m \delta_{u(m), u} [\overline{\theta_m} - \overline{\log \theta_m} - 1]$ \\
$\lambda_{0u}$         &$\sum_{t} N_{tu}$  &$\overline{\theta}_u H_{0u}\sum_t F_{tu}G_{tu}$ \\
$\lambda_{zuk}$         &$\sum_t N_{tu} \xi_{tk}$  &$\overline{\theta}_u H_{0u}\sum_t F_{tuk}G_{tu}$ \\
$\lambda_{xur}$         &$\sum_t N_{tu} x_{tr}$  &cf. Section \ref{non-conj} \\
$c_{0}$         &$U/2$  &$\sum_u\mathbb{E}_q \left[d_{0} \lambda_{0u} - \log \lambda_{0u} - \log d_{0} - 1\right]$ \\
$c_{zk}$         &$U/2$  &$\sum_u\mathbb{E}_q \left[d_{zk} \lambda_{zuk} - \log \lambda_{zuk} - \log d_{zk} - 1\right]$ \\
$c_{xr}$         &$U/2$  &$\sum_u\mathbb{E}_q \left[d_{xr} \lambda_{xur} - \log \lambda_{xur} - \log d_{xr} - 1\right]$ \\
$d_{0}$         &$U\mathbb{E}_q[c_{0}]$  &$\sum_u\mathbb{E}_q \left[c_{0}\lambda_{0u}\right]$ \\
$d_{zk}$         &$U\mathbb{E}_q[c_{zk}]$  &$\sum_u\mathbb{E}_q \left[c_{zk}\lambda_{zuk}\right]$ \\
$d_{xr}$         &$U\mathbb{E}_q[c_{xr}]$  &$\sum_u\mathbb{E}_q \left[c_{xr}\lambda_{xur}\right]$ \\
$\phi_{j u}$ for $j \in \{1, \ldots, \tau\}$  &$\sum_{\tau' \ge j} N_{\tau' u}$  &$\sum_{u, \tau \ge j} F_{t(\tau)} \prod_{\substack{\tau' \le \tau \\ \tau' \ne j}} \frac{\omega_{\tau' u}}{\zeta_{\tau' u}}$ \\
\end{tabular}
\end{center}
\end{table}

Here, we overload notation to write $N_{tu} = \sum_m \delta_{u(m), u} \delta_{t(m), t}\, N_m$, $\overline{\theta}_u = \sum_m \delta_{u(m), u}\overline{\theta}_m$, and make use of $H$, $G$, and $F$ as defined in Eq (\ref{HFG}) - (\ref{G}). Indeed, our implementation caches $F$ and $G$ for a substantial speedup (at the cost of additional memory requirements). For autocorrelated noise, the update rules are for the $j$-th item in the series of autocorrelation.

\subsection{Non-conjugate updates}
\label{non-conj}
We employ explicit optimization steps for two updates in our iterative Algorithm \ref{algo}. In each case, we employ an off-the-shelf optimization routine, though more efficient alternatives are likely possible.

\subsubsection{Covariate firing rate effects}
\label{beta_x}
Because we do not restrict the covariates $x(t)$ to be binary, Equation \ref{eff_rate} no longer yields a conditional log probability for $\lambda_x$ in the exponential family. This leaves us with a transcendental equation to solve for $\beta_{xr}$. However, from Table \ref{conj_updates}, we see that $\alpha_{xr} \gg \sum_t x_{tr}$, allowing us to approximate $G_t \approx \prod_r (\alpha_{xr}/\beta_{xr})^{x_{tr}}$ as we have done above. Moreover, since the sum $\sum_t G_t \sim T\overline{G}$, we expect $\alpha_x / \beta_x \approx 1$ at the optimum for most reasonable datasets. We thus reparameterize $\beta_{xur} = \alpha_{xur}e^{-\epsilon_{ur}}$ and write the relevant $\epsilon$-dependent piece of the objective function $\mathcal{L}$ as
\begin{equation}
    \mathcal{L}_\epsilon = \sum_{ur} \left[a_{xur}\epsilon_{ur} - b_{xur}e^{-\epsilon_{ur}} \right] - \sum_m \overline{\theta}_m H_{0u(m)} F_{t(m)u(m)}
    e^{-\sum_r \epsilon_{u(m)r} x_{t(m)u(m)r}}
\end{equation}
We also supply the optimizer with the gradient:
\begin{equation}
    \nabla_\epsilon \mathcal{L}_\epsilon = a_{xur} - b_{xur}e^{-\epsilon_{ur}}  - \sum_{m; u(m) = u} x_{t(m)ur} \overline{\theta}_m H_{0u} F_{t(m)u}
    e^{-\sum_r \epsilon_{ur} x_{t(m)ur}}
\end{equation}
where we sum only over observations with $u(m) = u$. On each iterate, we then optimize $\mathcal{L}_\epsilon$, initializing $\beta_x$ to the just-updated value of $\alpha_x$. In addition, we do not update firing rate effects for covariates separately, but optimize $\beta_{xur}$ for all $r$ together. Again, more efficient schemes are no doubt possible.

\subsubsection{Semi-Markov duration distribution}
\label{semimarkovdd}
If the dwell time distributions of states in the semi-Markov model were truly continuous, the parameters $(m, s^2)$ of a lognormal distribution $p(d|j)$ would have a Normal-Gamma conjugate prior, and updates would be closed-form. However, the requirement that the durations be integers and that there exist a maximal duration, $D$, over which these probability mass functions must normalize results in an extra term in $\mathbb{E}_q[\log p]$ arising from the normalization constant. In this case, we must explicitly optimize for the parameters of the Normal-Gamma prior on $(m, s^2)$. However, unlike the case of \ref{beta_x}, these parameters are only updated for one latent feature at a time. Since the Normal-Gamma distribution has only 4 parameters and the number of states of the latent variable is $S = 2$, this requires updating $4SK$ parameters, but only taken $4S = 8$ at a time, a much more manageable task.

To derive the optimization objective, we begin by noting that the semi-Markov model adds to the terms involving $\Xi$ and $\xi_0$ in latent state dynamics an additional piece:
\begin{multline}
    \sum_{d j k} C(d, j, k) \mathbb{E}_q \left[\log p_k(d|j) - \log \left(\sum_{d=1}^D p_k(d|j) \right) \right] \ge \\
    \sum_{d j k} C(d, j, k) \left[\mathbb{E}_q \left[\log p_k(d|j)\right] - \log \sum_{d=1}^D \mathbb{E}_q\left[p_k(d|j) \right] \right]
\end{multline}
where as noted above, the second term inside the expectation arises from the need to normalize $p(d|j)$ over discrete times and the inequality follows from Jensen's Inequality. Here, $d$ is the duration variable, $j$ labels states of the chain (here 0 and 1), and $k$, as elsewhere, labels chains\cite{Mitchell1993-sl,Mitchell1995-go,Yu2006-bb}. $C$ is defined for each state and chain as the probability of a dwell time $d$ in state $j$ \emph{conditioned on} the event of just having transitioned to $j$\footnote{Thus $C$ is equivalent to $\mathcal{D}_{t|T}$ in \cite{Yu2006-bb}.}
\footnote{
We also note that, just as the emission, transition, and initial state probabilities have their counterparts in variational parameters $(\eta, \tilde{A}, \tilde{\pi})$ in $q(z)$, so $C$ is matched by a term $-\sum_{djk} C(d, j, k) \nu_{djk}$ in $\mathbb{E}_q[-\log q]$. And in analogy with the HMM case, variation with respect to $\nu$ gives
\begin{equation}
    \nu_{djk} = \mathbb{E}_q \left[\log p_k(d|j)\right] - \log \sum_{d=1}^D \mathbb{E}_q\left[p_k(d|j) \right],
\end{equation}
implying that there are cancellations between $\log p(N, z|\Theta)$ and $\log q(z)$ in $\mathcal{L}$ and care must be taken when calculating it\cite{beal2003variational}.}

Thus the objective to be optimized is
\begin{equation}
    \sum_{d j k} C(d, j, k) \left[\mathbb{E}_q \left[\log p_k(d|j)\right] - \log \sum_{d=1}^D \mathbb{E}_q\left[p_k(d|j) \right] \right] + \mathbb{E}_q\left[\log \frac{p(m, \tau)}{q(m, \tau)} \right]
\end{equation}

For the case of $p(d|m, s^2)$ Log-Normal and $(m, \tau = s^{-2})$ Normal-Gamma, the terms $\mathbb{E}_q[\log p(d)]$ involve only routine expectations of natural parameters of the Normal-Gamma, and similarly for the expected log prior and entropy in the last term. Only slightly more complicated is the term arising from the normalization constant, which in the standard $(\mu, \lambda, \alpha, \beta)$ parameterization of the Normal-Gamma takes the form
\begin{equation}
    \mathbb{E}_q[p(d|m, \tau)] = \int d\tau dm \; p(d|i, m, \tau) q(m, \tau)
    = \frac{1}{\sqrt{2\pi}d} \sqrt{\frac{\lambda}{1 + \lambda}}
    \frac{\Gamma(\alpha + 1/2)}{\Gamma(\alpha)}
    \frac{\beta^{-\frac{1}{2}}}{\hat{\beta}^{\alpha + \frac{1}{2}}}
\end{equation}
with
\begin{equation}
    \hat{\beta} \equiv 1 + \frac{1}{2\beta} \frac{\lambda}{1 + \lambda}
    (\log d - \mu)^2
\end{equation}
This can be derived either by performing the integral directly or by noting that the result should be proportional to the posterior (Normal-Gamma, by conjugacy) of $(m, \tau)$ after having observed a single data point, $\log d$.

\begin{algorithm}[ht]
\caption{Iterative update for variational inference}\label{algo}
\begin{algorithmic}[1]
\Procedure{Iterate}{}
    \State Update baselines $\lambda_0$\Comment{conjugate Gamma}
    \State Update baseline hyperparameters $\gamma_0$\Comment{conjugate Gamma}
    \For{$k = 1\ldots K$}
        \State Update firing rate effects $\lambda_{zk}$\Comment{conjugate Gamma}
        \State Update firing rate hyperparameters $\gamma_{zk}$\Comment{conjugate Gamma}
        \State Calculate expected log evidence $\eta_k$
        \State Update Markov chain parameters $\tilde{A}_k, \tilde{\pi}_k$
        \State $\xi_k, \Xi_k, \log Z_k \gets$\Call{forward-backward}{$\eta_k, \tilde{A}_k, \tilde{\pi}_k$}
        \If{semi-Markov}
            \State Update duration distribution $p_k(d|j)$\Comment{BFGS optimization}
        \EndIf
        \State Update cached $F$
    \EndFor
    \State Update covariate firing effects $\lambda_x$\Comment{BFGS optimization}
    \State Update cached $G$
    \State Update overdispersion $\theta$\Comment{conjugate Gamma}
\EndProcedure
\end{algorithmic}
\end{algorithm}

\section{Experiments}
Code for all algorithms and analyses is available at \url{https://github.com/pearsonlab/spiketopics}.


\begin{flushleft}
{\Large
\textbf{Supplementary Information 2 \\ Categorization of multinomial logistic regression \\}
}
\end{flushleft}

\begin{figure}[!h]
    \includegraphics[width=.7\linewidth]{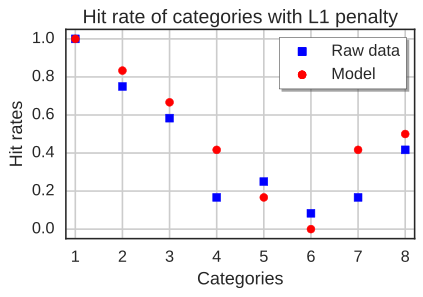}\\
    \includegraphics[width=.7\linewidth]{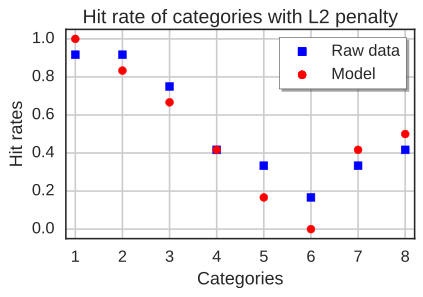}
	\caption{\bf Hit rates within eight categories} Compare the hit rates of actual data and inferred states of the macaque dataset. A. Multinomial logistic regression sing L1 regularization. Model inferred features outperform the raw data in 5/8 categories and tie in one category. Model wins. B. Multinomial logistic regression sing L2 regularization. Model inferred features outperform 3/8 categories and tie in one. Both tied.
	\label{fig:softmax}
\end{figure}

\nolinenumbers

%
%
%

\end{document}